\documentclass{article}
\usepackage{spconf,amsmath,graphicx}

\usepackage{lipsum}
\usepackage{booktabs}
\usepackage{graphicx}
\usepackage[normalem]{ulem}
\usepackage{hyperref}
\usepackage{multirow}
\useunder{\uline}{\ul}{}
\usepackage{caption}
\usepackage{subcaption}
\usepackage[subtle]{savetrees}
\usepackage{float}

\let\OLDthebibliography\thebibliography
\renewcommand\thebibliography[1]{
  \OLDthebibliography{#1}
  \setlength{\parskip}{0pt}
  \setlength{\itemsep}{0pt plus 0.3ex}
}

\usepackage{enumitem}
\setlist{nosep, leftmargin=14pt}

\usepackage{mwe} 

\title{Zero-Shot Pediatric Tuberculosis Detection in Chest X-Rays using Self-Supervised Learning}
\name{\parbox{\linewidth}{\centering
Daniel Capellán-Martín\textsuperscript{1,2},
Abhijeet Parida\textsuperscript{2},
Juan J. Gómez-Valverde\textsuperscript{1},
Ramon Sanchez-Jacob\textsuperscript{3},
Pooneh Roshanitabrizi\textsuperscript{2},
Marius G. Linguraru\textsuperscript{2,4},
María J. Ledesma-Carbayo\textsuperscript{1},
Syed M. Anwar\textsuperscript{2,4}
}}

\address{\parbox{\linewidth}{\centering\fontsize{9.5}{10}\selectfont
\textsuperscript{1} Biomedical Image Technologies, ETSI Telecomunicación, Universidad Politécnica de Madrid \& CIBER-BBN, Madrid, Spain \\
\textsuperscript{2} Sheikh Zayed Institute for Pediatric Surgical Innovation, Children’s National Hospital, Washington, DC, USA \\
\textsuperscript{3} Division of Pediatric Radiology, Children’s National Hospital, Washington, DC, USA \\
\textsuperscript{4} Departments of Radiology and Pediatrics, GWU School of Medicine, Washington, DC, USA
}}
\begin{document}
\ninept
\maketitle
\begin{abstract}
Tuberculosis (TB) remains a significant global health challenge, with pediatric cases posing a major concern. The World Health Organization (WHO) advocates for chest X-rays (CXRs) for TB screening. However, visual interpretation by radiologists can be subjective, time-consuming and prone to error, especially in pediatric TB. Artificial intelligence (AI)-driven computer-aided detection (CAD) tools, especially those utilizing deep learning, show promise in enhancing lung disease detection. However, challenges include data scarcity and lack of generalizability. In this context, we propose a novel self-supervised paradigm leveraging Vision Transformers (ViT) for improved TB detection in CXR, enabling zero-shot pediatric TB detection. We demonstrate improvements in TB detection performance ($\sim$12.7\% and $\sim$13.4\% top AUC/AUPR gains in adults and children, respectively) when conducting self-supervised pre-training when compared to fully-supervised (i.e., non pre-trained) ViT models, achieving top performances of 0.959 AUC and 0.962 AUPR in adult TB detection, and 0.697 AUC and 0.607 AUPR in zero-shot pediatric TB detection. As a result, this work demonstrates that self-supervised learning on adult CXRs effectively extends to challenging downstream tasks such as pediatric TB detection, where data are scarce.

\end{abstract}
\section{Introduction}
\label{sec:intro}

Tuberculosis (TB), caused by the bacterium \textit{Mycobacterium Tuberculosis}, remains a major global health problem, especially in developing regions, causing substantial mortality. Historically, TB outranked HIV/AIDS as the deadliest infectious disease until the emergence of COVID-19. However, the pandemic led to a decline in TB notifications, experiencing an increase in deaths due to reduced access to TB diagnosis and treatment \cite{WorldHealthOrganization2023GlobalReport}. Thus, timely detection and treatment are paramount to enhance patient outcomes and mitigate TB transmission. Moreover, the diagnosis and treatment of pediatric TB pose significant challenges, with a notable gap in the identification and treatment of affected children. Annually, 7.5 million children contract TB and 1.1 million develop the disease, contributing to 12\% of the global TB burden \cite{Jaganath2022TuberculosisChildren}. 
Effective screening is therefore essential to identify potential cases and facilitate referral for testing and preventive treatment among children \cite{Vonasek2021ScreeningChildren}.

The World Health Organization (WHO) advocates the use of chest X-rays (CXR) as a TB screening tool due to its widespread availability, cost-effectiveness, speedy results, and reduced risk of cross-infection \cite{WorldHealthOrganization2016ChestApproaches}. Moreover, in clinical practice, CXRs are the preferred imaging method for pediatric TB screening and diagnosis \cite{Ubaidi2018TheCare}. However, visual interpretation by radiologists in pediatric TB can be subjective, time-consuming, and error-prone. The challenges arise from non-specific symptoms and relatively fewer radiological manifestations in children compared to adults, leading to a low inter-radiologist agreement and potentially uncertain diagnoses \cite{Brady2017ErrorAvoidable, gomezvalverde2023}.

AI-driven computer-aided detection (CAD) tools, especially those using deep learning, hold promise for enhancing healthcare outcomes by improving radiologists' interpretation accuracy, reducing errors, and enabling patient-centric care. These tools show potential in addressing the diagnostic expertise gap in pediatric TB and facilitating large-scale TB screening through CXR image analysis, particularly in high TB burden areas \cite{Kulkarni2020ArtificialReview, Qin2021AImplementers, McBee2018DeepRadiology}. However, challenges include the scarcity of well-annotated medical data, especially in pediatric tasks, and a lack of generalizability across different clinical settings and devices. To overcome data scarcity (both in imaging data and in high-quality ground-truth), supervised and self-supervised pre-training strategies are employed. Self-supervised training allows models to learn meaningful features from unlabeled data, eliminating the need for costly and time-consuming annotations. Foundation models, such as those based on transformer architectures, have demonstrated remarkable performance across a wide range of natural language processing and computer vision tasks. These models, pre-trained on massive datasets, capture a broad understanding of the visual world, and fine-tuning them for specific medical imaging tasks presents an opportunity to leverage their robust generalization capabilities and tailor them for pediatric TB detection in CXRs.

Recently, multiple self-supervised learning strategies have demonstrated outstanding performance in natural image-based computer vision tasks, such as DINO \cite{Caron2021EmergingTransformers}, a knowledge distillation method without the need for labeled data, MoCo \cite{He2020MomentumLearning}, a momentum-based contrastive unsupervised learning method, and MAE \cite{He2022MaskedLearners}, based on masked autoencoders performing reconstruction tasks to learn meaningful representations. However, the application of these methods is still limited in the medical imaging domain, even more notably in the field of TB diagnosis.

In this context, we propose a novel self-supervised paradigm to address TB detection in CXRs for both adult and child patients. The main objective of our work is to propose a new self-supervised strategy with state-of-the-art performance on adult and pediatric TB detection requiring no supervised pre-training and minimal fine-tuning. We outline our contributions as follows:

\begin{enumerate}
  \item We propose a novel self-supervised pre-training paradigm aimed at addressing pediatric TB detection in chest X-ray imaging using a ViT backbone.
  \item We conduct comprehensive domain-specific downstream tasks on CXR images, using multiple self-supervised methods, both on adult and pediatric TB data, from public and private sources, to demonstrate the importance of pre-training on domain data for improved performance.
  \item To the best of our knowledge, we are the first to (i) demonstrate that self-supervised pre-training on adult chest X-rays (CXRs) contributes to learning valuable representations for TB detection in both adult and pediatric CXRs, with the latter being underrepresented in most datasets and more difficult to detect, and (ii) explore zero-shot classification in pediatric TB.
\end{enumerate}

\section{Related Work}
\label{sec:rel-work}

Several studies have explored deep learning (DL) techniques for TB detection in CXRs. Some approaches include CNNs and transfer learning techniques \cite{hwang2016novel, Islam2017AbnormalityNetworks}. Additional methods have explored ResNet-based solutions, emphasizing not only TB detection performance, but also efficiency and low computational requirements \cite{Pasa2019EfficientVisualization, Capellan-Martin2023ADetection}). Some studies have also examined DenseNet-121 models for improving diagnosis in HIV-infected TB patients \cite{Rajpurkar2020CheXaid:HIV}, while others have studied self-attention mechanisms \cite{Wong2022TB-Net:Images}. However, current approaches exhibit limited performance when tested on pediatric patients, with the domain of research in this specific age group remaining considerably underexplored. This underscores the urgent need to create AI systems tailored for infants and children. Recently, Palmer et al.\@ \cite{Palmer2023OptimisingChildren} reported, for the first-time ever, results of a CAD software (CAD4TB) applied on pediatric CXRs to identify TB. Notably, they reported an AUC of 0.58 with a fully-supervised non-fine-tuned model, and an AUC of 0.72 with a fully-supervised fine-tuned model. Finally, although limited, some recent studies \cite{sowrirajan2021moco,gazda2021,Park2022Self-evolvingDistillation,Tiu2022Expert-levelLearning,Anwar2023SPCXR:Model} verse about the application of self-supervised techniques to improve the detection of diverse lung diseases on CXR images, moving towards the concept of foundation medical image analysis models.

\section{Methods}
\label{sec:methods}

\subsection{Datasets}
\label{sec:datasets}

During the process of self-supervised pre-training, a large-scale dataset of 357,286 frontal (antero-posterior (AP) and postero-anterior (PA)) CXR images was used, sourced from diverse public CXR datasets:
\begin{itemize}
    \item CheXpert \cite{chexpert2019} (191,229 frontal CXR images from 64,740 patients were used) and ChestXray-14 \cite{Wang2017ChestX-ray8:Diseases} (112,120 frontal CXR images from 30,805 patients were used), which conformed the majority of the pre-training data. Both publicly available datasets are weakly-annotated using natural language processing (NLP) techniques, yielding $>$90\% labeling accuracy. They encompass annotations for the presence of 14 observations as positive, negative, or uncertain. Examples of radiological observations present in both datasets include lung opacity, atelectasis, pneumonia, cardiomegaly and pneumothorax, among others.
    \item The COVIDx CXR-3 dataset\footnote{\href{https://www.kaggle.com/datasets/andyczhao/covidx-cxr2}{https://www.kaggle.com/datasets/andyczhao/covidx-cxr2}} \cite{Wang2020COVID-Net:Images, pavlova2022covidx}. It is the most extensive and diverse COVID-19 CXR dataset in open access form, comprising 30,386 frontal CXR images taken from 17,026 patients.
    \item The SPR X-Ray Gender Prediction Challenge\footnote{\href{https://kaggle.com/competitions/spr-x-ray-gender}{https://kaggle.com/competitions/spr-x-ray-gender}}, intended to determine gender and age using CXRs. A total of 22,449 images were incorporated from this source.
    \item Shenzhen Hospital X-ray Set (SZ) \cite{Jaeger2014TwoDiseases}, Belarus (BL)\footnote{\href{https://github.com/frapa/tbcnn}{https://github.com/frapa/tbcnn}}, and Montgomery County X-ray Set (MC) \cite{Jaeger2014TwoDiseases} datasets, intended for supporting TB research. A total of 1,102 frontal CXR images were incorporated from these three datasets (662, 302 and 138 from SZ, BL and MC, respectively).
\end{itemize}

\noindent
During the fine-tuning and evaluation phases, we used both adult and pediatric cohorts. For the adult experiments, SZ and MC were used (details specified above). Adult data for fine-tuning were split into training (80\%, 640 frontal CXR images) and independent test (20\%, 160 frontal CXR images, no patient overlap). Stratification was applied based on cohort, label, sex and age. For pediatrics, we used an in-house collected dataset from Children's National Hospital (CNH), Washington, DC, USA, featuring 85 patients aged 0-18 years (32.94\% 0-3 years, 29.41\% 4-12 years, 37.65\% 13-18 years). A total of 85 frontal CXR images were used (54 negative, 31 positive), configuring an unseen, out-of-domain (ODD), independent test set to evaluate, in a zero-shot fashion, our method.

MC and SZ standard TB references were either confirmed microbiologically or, when unfeasible, based on consistent clinical symptoms and imaging findings. On the other hand, CNH reference TB standards were confirmed microbiologically, except for one unconfirmed TB case. Among the 85 patients, three had a latent TB infection.

\subsection{Architecture and self-supervised strategy}
\label{sec:vit-ssl}

Vision transformers (ViT) \cite{Dosovitskiy2020AnScale} have recently shown great success in computer vision tasks, and their use in the medical domain is growing at an unprecedented rate. The ViT processes input images by tokenizing them into a sequence of patches. Each patch, obtained from the image's height, width, and channels ($H \times W \times C$), is flattened into $n$ two-dimensional patches of size $p_1 \times p_2 \times C$ pixels. A linear layer or a convolutional layer with specific parameters projects these patches to $D$ hidden dimensions. To preserve spatial relationships, fixed or learnable position embeddings are added to patch embeddings. The Transformer encoder comprises $L$ consecutive Multi-head Self-Attention (MSA) and Multi-Layer Perceptron (MLP) blocks. The MSA layer employs a trainable triplet of (query (Q), key (K), value (V)), producing output sequences that are linearly projected to $n \times D$ dimensions. The self-attention mechanism is based on the equation \ref{eq:att}, where $d_{k}$ represents the dimensionality of the key vectors in the attention mechanism. For classification, a trainable class token is added to the patch token sequence, processed through the Transformer encoder, and connected to a classification head. However, in unsupervised scenarios, such as self-supervised (SSL) learning, no classification token is required. 

\begin{equation}
    {\rm Attention}(\mathbf{Q}, \mathbf{K}, \mathbf{V}) = {\rm softmax}(\frac{\mathbf{Q}\mathbf{K}^T}{\sqrt{d_{k}}})\mathbf{V}
    \label{eq:att}
\end{equation}

For all the experiments, we used a vision transformer (ViT) \cite{Dosovitskiy2020AnScale} (ViT-S/16) as the backbone. During pre-training, we incorporated diverse SSL methods, including MAE \cite{He2022MaskedLearners}, MOCO-v3 \cite{He2020MomentumLearning, Chen2021AnTransformers} and DINO \cite{Caron2021EmergingTransformers}. Also, fully-supervised, i.e., non pre-trained, ViT models were also used for baseline comparison. In contrastive SSL methods, instead of using classification tokens, we use a contrastive token and compute contrastive losses. On the other hand, in image reconstruction methods, we usually partially mask patches of the input (i.e., data tokens) and perform image reconstruction tasks. After pre-training, we conducted both linear probing and fine-tuning. During linear probing, the pre-trained model's weights are frozen, except for the weights of the classification head, which is then trained or "probed" to adapt the model to the specifics of the new task. During fine-tuning, instead, the entire pre-trained model layers are unfrozen, and the model is trained on the specifics of the new task. Fine-tuning is more intensive computationally compared to linear probing, but allows the model to adjust more parameters to better fit the new task.

\begin{figure*}[htbp]
    \centering
    \includegraphics[width=0.975\textwidth]{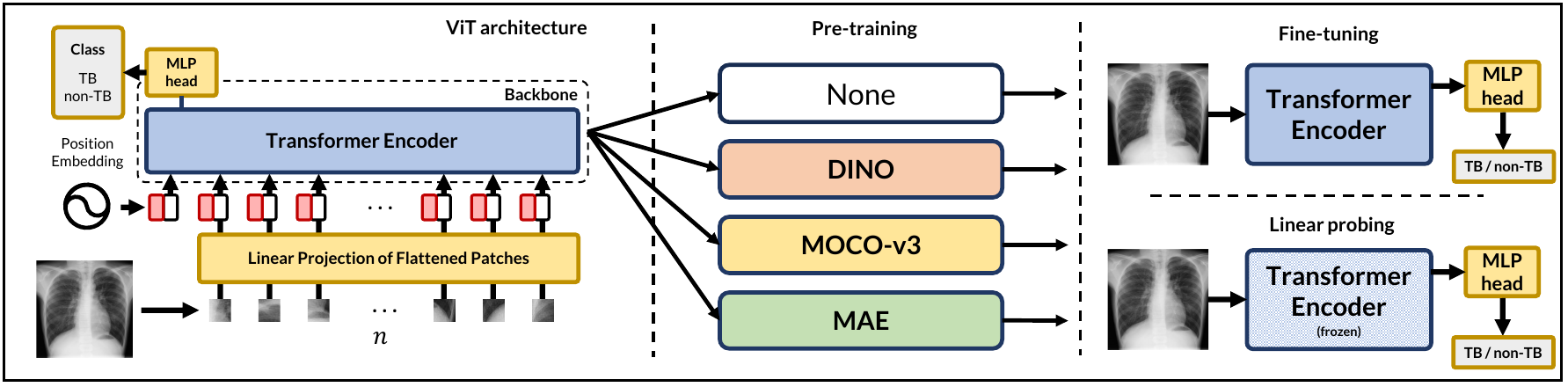}
    \vspace*{-0.2cm}
    \caption{We propose a novel self-supervised strategy leveraging ViTs for improved TB detection in CXR, enabling for zero-shot pediatric TB detection.\vspace{-0.3cm}}
    \label{fig:pipeline}
\end{figure*}

Consequently, the self-supervised learning strategy proposed in this work is illustrated in Figure \ref{fig:pipeline}. It encompasses multiple steps described in this work, spanning from pre-training to TB prediction.

\section{Results}
\label{sec:exp-results}

\subsection{Experimental Setup}
\label{sec:exp-setup}

We conducted several downstream classification experiments to evaluate our approach on both adult (seen distribution) and pediatric (unseen, out-of-domain (OOD) and shifted distribution) TB datasets, employing diverse self-supervised learning methods (see Section \ref{sec:vit-ssl}). We executed two downstream binary (non-TB vs.~TB) classification tasks. One task utilized an independent subset of adult CXR data (MC \& SZ), with no distribution shift, i.e., similar data distribution as the one used for fine-tuning. The second task, following a zero-shot approach, involved an independent set of pediatric CXR data (CNH). This dataset presented distribution shifts due to distinct acquisition devices and setups, along with differences in age with respect to the fine-tuning datasets.

During the pre-training phase, regardless of the self-supervised approach, we maintained the following set-up: batch size of 64, 200 epochs, cosine learning rate (LR) scheduler ($LR_{min} = 1e^{-6}$, $LR_{initial} = 5e^{-4}$, 10 warm-up epochs), cosine weight decay scheduler (weight decay of 0.04) and AdamW optimizer \cite{Loshchilov2017DecoupledRegularization}. Subsequently, we fine-tuned the pre-trained models (also non-pretrained, for the sake of comparison) using a batch size of 48, 200 epochs and the same LR scheduler settings as in the pre-training. Pre-training was performed across multiple decentralized workstations utilizing multi-GPU settings. The fine-tuning step, however, was executed on the Magerit-3 supercomputer from Universidad Politécnica de Madrid, Spain, dedicating 1 NVIDIA V100/A100 GPU per experiment.

In all experiments, we employed a vision transformer \cite{Dosovitskiy2020AnScale} (ViT-S/16) as the backbone, and an image size of 256$\times$256. During evaluation, confidence intervals (CI) were calculated using DeLong's approach \cite{delong, delong-fast} for area under the receiver operating characteristic (ROC) curve (AUC) metrics and bootstrap method \cite{bootstrap} (2000 repetitions) for accuracy (ACC) and area under the precision-recall curve (AUPR).

\subsection{TB classification results}
\label{sec:results}

\subsubsection{Evaluation on adult data}
\label{sec:adult-results}

We conducted linear probing and fine-tuning experiments on adult TB data, i.e., using the independent test subsets of the Shenzhen Hospital X-ray Set (SZ) and the Montgomery County X-ray Set (MC). The results, as reported in Table \ref{tab:results-mcsz}, demonstrated that MAE achieved the best performance, achieving the highest AUC and AUPR scores: 0.926 and 0.932, respectively. This occurred during linear probing on the SZ \& MC training subsets. DINO achieved the second-best AUPR and AUC scores and the best ACC and sensitivity (SN) values. On the other hand, when conducting fine-tuning on the same training subset, MOCO-v3 exhibited the highest AUC score (0.959 AUC), and DINO exhibited the highest AUPR score (0.962 AUPR). DINO also scored the highest ACC and SN values. We can also notice a top performance gain of $\sim$12.7\% and $\sim$11.3\% in AUC and AUPR, respectively, for linear probing, and $\sim$10.7\% and $\sim$11.3\% in AUC and AUPR, respectively, for fine-tuning, when compared to a fully-supervised, i.e., non pre-trained, model performance. Additionally, the highest AUC and AUPR scores were attained through fine-tuning rather than linear probing. The ROC curves for the fine-tuned models when tested on MC \& SZ are displayed in Figure \ref{fig:roc-mcsz}.

\subsubsection{Evaluation on unseen pediatric data}
\label{sec:ped-results}

\begin{figure*}[htbp]
     \centering
     \begin{subfigure}[b]{0.4\textwidth}
         \centering
         \includegraphics[width=\textwidth]{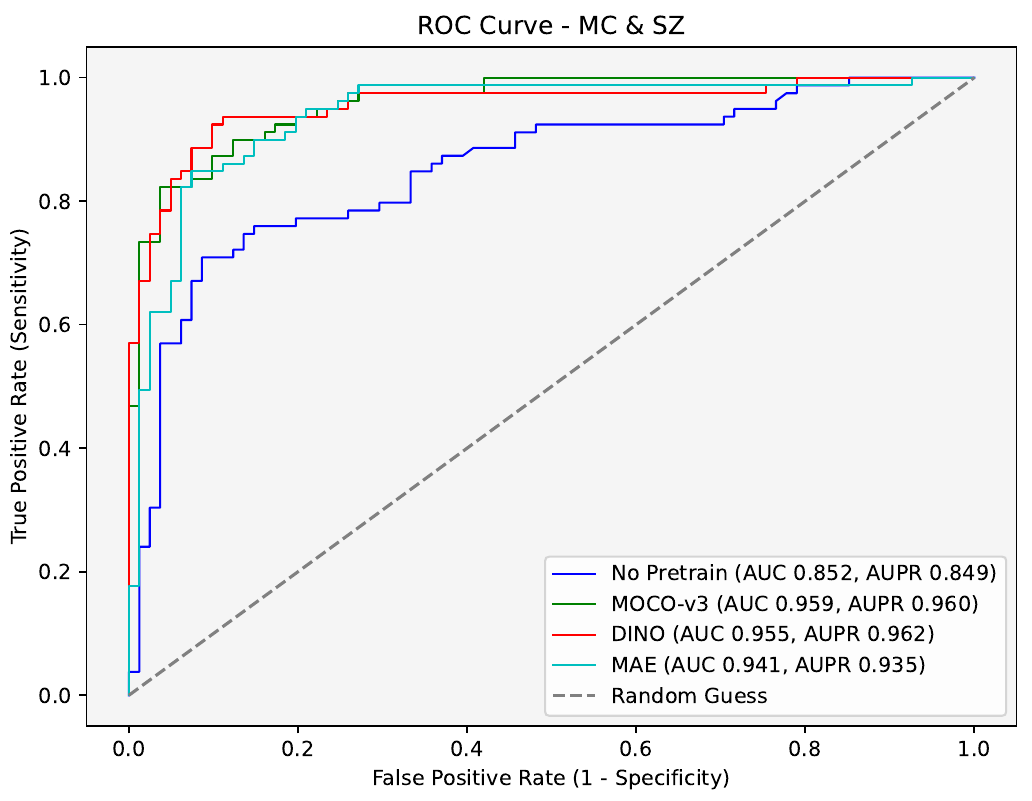}
         \vspace{-0.5cm}
         \caption{}
         \label{fig:roc-mcsz}
     \end{subfigure}
     \hspace{0.1\textwidth}
     \begin{subfigure}[b]{0.4\textwidth}
         \centering
         \includegraphics[width=\textwidth]{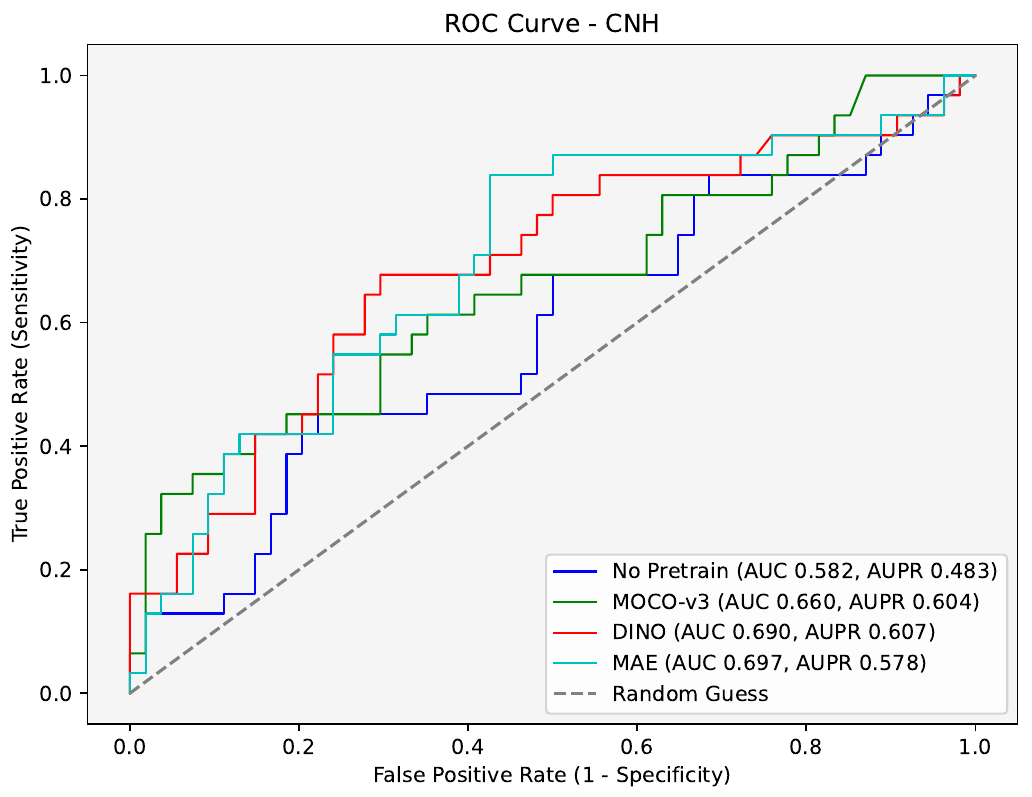}
         \vspace{-0.5cm}
         \caption{}
         \label{fig:roc-cnh}
     \end{subfigure}
     \vspace{-0.7\baselineskip}
    \caption{Receiver operating characteristic (ROC) curves when conducting fine-tuning on adult TB data (MC \& SZ training subsets) and: (a) testing on an independent adult TB dataset; (b) performing zero-shot inference on an independent, out-of-domain pediatric TB dataset. \vspace{-0.5cm}}
    \label{fig:three graphs}
\end{figure*}

With the models obtained in the previous experiments, i.e., those resulting from conducting linear probing and fine-tuning on adults (MC \& SZ), we performed zero-shot classification on an unseen, out-of-domain, i.e., distribution-shifted, pediatric TB dataset collected in-house at Children's National Hospital (CNH). This enables for assessing transferability of local and global features and patterns learnt from CXRs during pre-training and transfer learning. The results are reported in Table \ref{tab:results-cnh}. In this other scenario, when conducting linear probing on adults (MC \& SZ) and zero-shot testing on children, DINO demonstrated a superior performance (0.622 AUC, 0.560 AUPR) when compared to alternative self-supervised methods and a fully-supervised approach. On the other hand, when instead conducting fine-tuning on adults (MC \& SZ) and zero-shot testing on children, MAE achieved a higher AUC score (0.697 AUC), and DINO a higher AUPR score (0.607), when compared to alternative self-supervised methods and a fully-supervised model. The highest ACC, SN and specificity (SP) values were achieved with MOCO-v3. Additionally, we were able to achieve a top performance improvement of $\sim$10\% and $\sim$13.4\% in AUC and AUPR, respectively, for linear probing, and $\sim$11.5\% and $\sim$12.4\% in AUC and AUPR, respectively, for fine-tuning, when compared to a fully-supervised model. As in the previous experiments, the highest AUC and AUPR scores were obtained through fine-tuning rather than linear probing. As a result, the ROC curves for the fine-tuned models during zero-shot pediatric TB testing are depicted in Figure \ref{fig:roc-cnh}.

\begin{table*}[htbp]
\centering
\setlength{\tabcolsep}{3pt}
\setlength\doublerulesep{0.8pt}
\resizebox{\textwidth}{!}{%
\begin{tabular}{@{}lcccccccccc@{}}
\toprule
\multicolumn{1}{c}{\multirow{2}{*}{\textbf{\begin{tabular}[c]{@{}c@{}} \\[-2.5mm] SSL \\ Technique\end{tabular}}}} & \multicolumn{5}{c}{\textbf{Linear Probing}} & \multicolumn{5}{c}{\textbf{Fine-tuning}} \\ \cmidrule(l){2-11} 
\multicolumn{1}{c}{} & \textbf{ACC (95\% CI)} & \textbf{AUPR (95\% CI)} & \textbf{AUC (95\% CI)} & \textbf{TPR (SN)} & \multicolumn{1}{c|}{\textbf{TNR (SP)}} & \textbf{ACC (95\% CI)} & \textbf{AUPR (95\% CI)} & \textbf{AUC (95\% CI)} & \textbf{TPR (SN)} & \textbf{TNR (SP)} \\ \toprule[0.8pt]\midrule[0.3pt]
\multicolumn{1}{l|}{\textbf{No Pretrain}} & .744 (.675, .812) & .819 (.739, .892) & .799 (.729, .869) & .620 & \multicolumn{1}{c|}{.864} & .794 (.725, .856) & .849 (.762, .926) & .852 (.792, .913) & .696 & .889 \\ \midrule
\multicolumn{1}{l|}{\textbf{MOCO-v3}} & .731 (.656, .8) & .823 (.734, .895) & .81 (.743, .877) & .671 & \multicolumn{1}{c|}{.790} & .881 (.825, .931) & {\ul .96 (.928, .984)} & \textbf{.959 (.933, .985)} & .823 & \textbf{.938} \\
\multicolumn{1}{l|}{\textbf{DINO}} & \textbf{.862 (.806, .912)} & {\ul .925 (.875, .962)} & {\ul .919 (.878, .961)} & \textbf{.848} & \multicolumn{1}{c|}{{\ul .877}} & \textbf{.9 (.85, .944)} & \textbf{.962 (.931, .986)} & {\ul .955 (.923, .987)} & \textbf{.873} & {\ul .926} \\
\multicolumn{1}{l|}{\textbf{MAE}} & {\ul .85 (.794, .906)} & \textbf{.932 (.887, .966)} & \textbf{.926 (.886, .966)} & {\ul .810} & \multicolumn{1}{c|}{\textbf{.889}} & {\ul .888 (.838, .931)} & .935 (.879, .975) & .941 (.903, .979) & {\ul .848} & {\ul .926} \\ \midrule
\multicolumn{11}{l}{\begin{tabular}[c]{@{}l@{}}ACC, accuracy; AUPR, Area Under the Precision-Recall (PR) Curve; AUC, Area Under the Receiver Operating Characteristic (ROC) Curve; \\ TPR, True Positive Rate; SN, Sensitivity; TNR, True Negative Rate; SP, Specificity; CI: Confidence Interval.\end{tabular}} \\ \bottomrule
\end{tabular}%
}
\vspace{-0.25cm}
\caption{\textbf{Adult TB classification results} obtained from linear probing and fine-tuning on an independent adult test set (seen distribution) from \textbf{Montgomery (MC) \& Shenzhen (SZ)} datasets. The highest scores are highlighted in bold, and the second best ones are underlined.\vspace{-0.15cm}}
\label{tab:results-mcsz}
\end{table*}

\begin{table*}[htbp]
\centering
\setlength{\tabcolsep}{3pt}
\setlength\doublerulesep{0.8pt}
\resizebox{\textwidth}{!}{%
\begin{tabular}{@{}lcccccccccc@{}}
\toprule
\multicolumn{1}{c}{\multirow{2}{*}{\textbf{\begin{tabular}[c]{@{}c@{}} \\[-2.5mm] SSL \\ Technique\end{tabular}}}} & \multicolumn{5}{c}{\textbf{Linear Probing}} & \multicolumn{5}{c}{\textbf{Fine-tuning}} \\ \cmidrule(l){2-11} 
\multicolumn{1}{c}{} & \textbf{ACC (95\% CI)} & \textbf{AUPR (95\% CI)} & \textbf{AUC (95\% CI)} & \textbf{TPR (SN)} & \multicolumn{1}{c|}{\textbf{TNR (SP)}} & \textbf{ACC (95\% CI)} & \textbf{AUPR (95\% CI)} & \textbf{AUC (95\% CI)} & \textbf{TPR (SN)} & \textbf{TNR (SP)} \\ \toprule[0.8pt]\midrule[0.3pt]
\multicolumn{1}{l|}{\textbf{NoPretrain}} & .588 (.482, .694) & .426 (.296, .609) & .522 (.384, .659) & \textbf{.484} & \multicolumn{1}{c|}{.648} & .659 (.541, .753) & .483 (.337, .658) & .582 (.451, .712) & \textbf{.355} & .833 \\ \midrule
\multicolumn{1}{l|}{\textbf{MOCO-v3}} & .529 (.424, .635) & .351 (.255, .516) & .478 (.351, .604) & {\ul .258} & \multicolumn{1}{c|}{.685} & \textbf{.718 (.624, .812)} & {\ul .604 (.439, .767)} & .66 (.534, .786) & \textbf{.355} & \textbf{.926} \\
\multicolumn{1}{l|}{\textbf{DINO}} & \textbf{.682 (.576, .776)} & \textbf{.56 (.404, .732)} & \textbf{.622 (.491, .753)} & .194 & \multicolumn{1}{c|}{\textbf{.963}} & .659 (.553, .765) & \textbf{.607 (.446, .765)} & {\ul .69 (.567, .812)} & .226 & {\ul .907} \\
\multicolumn{1}{l|}{\textbf{MAE}} & {\ul .612 (.506, .718)} & {\ul .439 (.311, .614)} & {\ul .55 (.42, .68)} & .129 & \multicolumn{1}{c|}{{\ul .889}} & {\ul .682 (.576, .788)} & .578 (.424, .763) & \textbf{.697 (.577, .817)} & .323 & .889 \\ \midrule
\multicolumn{11}{l}{\begin{tabular}[c]{@{}l@{}}ACC, accuracy; AUPR, Area Under the Precision-Recall (PR) Curve; AUC, Area Under the Receiver Operating Characteristic (ROC) Curve; \\ TPR, True Positive Rate; SN, Sensitivity; TNR, True Negative Rate; SP, Specificity; CI: Confidence Interval.\end{tabular}} \\ \bottomrule
\end{tabular}%
}
\vspace{-0.25cm}
\caption{\textbf{Zero-shot pediatric TB classification results} obtained from linear probing and fine-tuning on adults (MC \& SZ) and tested on CNH dataset (independent, unseen, ODD, shifted distribution). The highest scores are highlighted in bold, and the second best ones are underlined.\vspace{-0.15cm}}
\label{tab:results-cnh}
\end{table*}

\section{Discussion}
\label{sec:discussion}

The conclusions drawn from the downstream TB classification experiments reveal that self-supervised pre-training on a ViT model leads to superior TB detection performance compared to fully-supervised alternatives. Depending on our goals, a range of self-supervised methods can be employed to enhance TB detection. Method comparisons can be conducted focusing on diverse metrics, although caution should be exercised when comparing methods using metrics such as accuracy due to potential bias from class imbalance. Fine-tuning/linear probing on children may increase SN and SP values, requiring more data. SN and SP values may vary with different prediction thresholds, making AUC and AUPR better for unbiased model performance assessment and results comparison, with AUPR being particularly advantageous for tasks involving class imbalance. As a result, when focusing on AUPR, DINO achieved better pediatric (out-of-distribution, unseen data) TB detection performance, while MAE scored a top AUC of 0.697, aligning closely with Palmer et al.'s \cite{Palmer2023OptimisingChildren} pediatric TB performance (0.72 AUC) and surpassing previous zero-shot results (0.58 AUC) in the same study. These findings demonstrate efficient knowledge transfer to an out-of-distribution scenario, i.e., pediatric TB, where radiological signs and disease detection pose unique challenges in clinical practice.

It’s important to note that using MC and SZ datasets for the fine-tuning and linear probing steps, as part of a seen distribution experiment, does not imply overfitting during inference. These self-supervised
tasks induce learning of general features from given CXR distribution, rather than task-specific patterns. In addition, experiments were conducted on an unseen OOD dataset to assess the generalizability of the model to different scenarios.

These findings demonstrate the efficacy of self-supervised pre-training with a ViT backbone in enhancing TB detection performance compared to fully-supervised approaches. This emphasizes the effectiveness of foundational models in addressing challenges like data scarcity, especially in tasks such as pediatric TB.

\vspace{-0.25cm}
\section{Conclusion}
\label{sec:conclusion}
\vspace{-0.1cm}

Given the increasing emphasis on CAD systems for TB, particularly in pediatric contexts, we have proposed a novel self-supervised paradigm to tackle complex downstream tasks like TB detection. We demonstrated how the transfer of rich learning representations learned from CXRs enhances TB detection performance, not only in adult TB but also when conducting zero-shot classification on pediatric TB. This research contributes to advancing the application of foundational models in challenging pediatric diagnostic tasks with limited data, aligning with the goal of improving clinical care and patient outcomes in TB diagnosis.

\vspace{6pt}
\noindent
\textbf{Compliance with ethical standards} ~ This research study was conducted using retrospective data from both open access and non-public sources (details in text). Ethical approval was granted from the Institutional Review Board (IRB) from the corresponding institutions.

\vspace{4pt}
\noindent
\textbf{Acknowledgements} ~ This work was supported by H2020-MSCA-RISE-2018 INNOVA4TB (EU) project (ID 823854), ADVANCE-TB Cost Action (CA21164 (EU)) and by Ministerio de Ciencia e Innovación, Agencia Estatal de Investigación grants PDC2022-133865-I00 and PID2022-141493OB-I0 (MCIN/AEI/10.13039/501100011033). The authors gratefully acknowledge the Universidad Politécnica de Madrid for providing computing resources on Magerit Supercomputer. We acknowledge Taylor Broudy for her participation in clinical data extraction.
\vspace{-0.2cm}

\bibliographystyle{IEEEbib-abbrev-etal}
\bibliography{refs_manual}

\end{document}